\documentclass[10pt]{article} 
\usepackage[preprint]{tmlr}
\usepackage{amsfonts}
\usepackage{amssymb}

\usepackage{amsmath,amsfonts,bm}

\def\eqref#1{equation~\ref{#1}}

\def\1{\bm{1}}

\DeclareMathAlphabet{\mathsfit}{\encodingdefault}{\sfdefault}{m}{sl}
\SetMathAlphabet{\mathsfit}{bold}{\encodingdefault}{\sfdefault}{bx}{n}

\usepackage{hyperref}
\usepackage{url}
\usepackage{enumitem}
\usepackage{amsthm}
\usepackage{graphicx}
\usepackage{booktabs}
\usepackage{subcaption}
\newtheorem{theorem}{Theorem}[section]        
\newtheorem{lemma}[theorem]{Lemma}            

\newtheorem{proposition}[theorem]{Proposition}
\newcommand\blfootnote[1]{%
  \begingroup
  \renewcommand\thefootnote{}
  \footnotetext{#1}
  \endgroup
}

\begin{document}

\theoremstyle{definition}
\newtheorem{definition}[theorem]{Definition}
\newtheorem{remark}[theorem]{Remark}

\title{Symbolic Feedforward Networks for Probabilistic Finite \\Automata: Exact Simulation and Learnability}


\author{\name Sahil Rajesh Dhayalkar \email sdhayalk@asu.edu \\
      \addr Arizona State University}


\newcommand{\fix}{\marginpar{FIX}}
\newcommand{\new}{\marginpar{NEW}}

\def\month{MM}  
\def\year{YYYY} 
\def\openreview{\url{https://openreview.net/forum?id=XXXX}} 

\maketitle

\blfootnote{This paper has been submitted to IEEE Access and is currently under review.}

\begin{abstract}
We present a formal and constructive theory showing that probabilistic finite automata can be exactly simulated using symbolic feedforward neural networks. Our architecture represents state distributions as vectors and transitions as stochastic matrices, enabling probabilistic state propagation via matrix-vector products. This yields a parallel, interpretable, and differentiable simulation of probabilistic finite automata dynamics using soft updates without recurrence. We formally characterize probabilistic subset construction, finite $\varepsilon$-closure, and exact simulation via layered symbolic computation, and prove equivalence between probabilistic finite automata and specific classes of neural networks. We further show that these symbolic simulators are not only expressive but learnable: trained with standard gradient descent-based optimization on labeled sequence data, they recover the exact behavior of ground-truth probabilistic finite automata. This learnability, formalized in Proposition~\ref{PROP:LEARNABILITY}, is the crux of this work. Our results unify probabilistic automata theory with classical neural architectures under a rigorous algebraic framework, bridging the gap between symbolic computation and deep learning.
\end{abstract}

\section{Introduction}
\label{introduction}

Neural networks and automata represent two foundational paradigms in machine learning and theoretical computer science. Neural networks are continuous, differentiable, and data-driven; finite automata are discrete, symbolic, and rule-based. Bridging these paradigms has long been a goal of neural-symbolic learning~\cite{besold2017neuralsymboliclearningreasoningsurvey, garcez2019neuralsymboliccomputingeffectivemethodology}, but most prior attempts have been empirical or heuristic, lacking rigorous guarantees of simulation or interpretability. This paper presents the first complete theoretical and empirical framework that constructs, simulates, and learns \textit{probabilistic finite automata} (PFAs) using \textit{symbolic feedforward neural networks}.

We focus on PFAs—a powerful generalization of deterministic and nondeterministic automata—which model the stochastic evolution of state under probabilistic transitions~\cite{RABIN1963230, Paz1971IntroductionTP}. Unlike deterministic finite automata (DFAs), where each input induces a unique state trajectory, PFAs compute the \textit{acceptance probability} of a string by marginalizing over exponentially many paths, making them expressive yet computationally delicate. Despite their theoretical richness, PFAs remain largely unexplored in the context of modern neural networks. Existing works either ignore stochasticity or simulate automata using black-box recurrent models with limited interpretability~\cite{weiss2018practical, hupkes2020compositionalitydecomposedneuralnetworks, adaAX}.

This work closes that gap. We show that feedforward networks—without recurrence, memory, or sampling—can simulate PFAs exactly. Each layer in our architecture implements a stochastic state update using matrix-vector multiplication with symbolic transition matrices, followed by a softmax or sigmoid operation for normalization and classification. Our framework extends prior work on DFA~\cite{dhayalkar2025neuralnetworksuniversalfinitestate} and NFA~\cite{dhayalkar2025nfa} simulation by unifying deterministic, nondeterministic, and probabilistic transitions under a single symbolic neural model.

Formally, we define the simulation of PFAs by feedforward networks as the exact computation of the acceptance probability for any string via symbolic matrix composition. Proposition~\ref{PROP:SIMULATION_OF_PFA} and Theorem~\ref{THM:EQUIVALENCE} prove a bidirectional equivalence: (1) every PFA can be simulated by a depth-unrolled feedforward network with shared stochastic parameters, and (2) every such symbolic network corresponds to a valid PFA. This allows us to replace automaton-based reasoning with static, interpretable network computation. Moreover, we define new symbolic mechanisms for probabilistic $\varepsilon$-closure (Lemma~\ref{LEM:PROBABILISTIC_EPSILON_CLOSURE}), probabilistic subset construction (Theorem~\ref{THM:PROBABILISTIC_SUBSET_CONSTRUCTION}), and belief state propagation (Proposition~\ref{PROP:PROBABILISTIC_STATE_VECTOR_REPRESENTATION}), all implemented as differentiable neural layers.

Critically, our framework also supports \textbf{learnability}, which is the crux of our paper. Proposition~\ref{PROP:LEARNABILITY} shows that symbolic feedforward networks not only simulate known PFAs but can be trained via gradient descent techniques to exactly recover unknown PFAs from data. We construct a learnable architecture where transition matrices are parameterized and normalized to remain stochastic. When trained with probabilistic acceptance labels, the network converges to exact simulators of the ground-truth PFA—achieving perfect accuracy and full semantic fidelity. This distinguishes our work from prior approaches that treat networks as black-box approximators without structure~\cite{butoi2025trainingneuralnetworksrecognizers}, and offers a new paradigm: training symbolic automata directly within neural networks.

Extensive experiments validate each theoretical result. We empirically confirm that our symbolic state vectors preserve probabilistic semantics, match ground-truth PFA dynamics exactly at each timestep, and simulate full acceptance behavior across randomly generated PFAs. Our learned models achieve exact recovery of the underlying automaton, demonstrating both expressivity and interpretability.

The key contributions of this work are as follows.
\begin{itemize}
    \item We present the first constructive theory for simulating probabilistic finite automata (PFAs) with $\varepsilon$-transitions using symbolic feedforward networks.
    \item We introduce exact formulations of probabilistic subset construction and finite $\varepsilon$-closure via matrix-vector dynamics, enabling differentiable automaton emulation.
    \item We prove bidirectional equivalence between feedforward networks and PFAs (Theorem~\ref{THM:EQUIVALENCE}), establishing a formal link between symbolic state machines and neural models.
    \item We demonstrate \textbf{learnability} (Proposition~\ref{PROP:LEARNABILITY}): symbolic neural networks can be trained from data to match ground-truth PFAs exactly, offering both statistical and structural fidelity. This is the central result of this paper.
    \item We validate each theoretical claim through rigorous experiments across multiple random seeds, with full statistical reporting and semantic evaluations.
\end{itemize}

This work advances the field of neural-symbolic computation by providing a unified, interpretable, and learnable framework for simulating probabilistic automata in neural networks. It lays the groundwork for deeper integration of formal language theory and deep learning, with applications in structured reasoning, interpretable sequence modeling, and symbolic control. 

\section{Related Work}
\label{related_work}
The study of neural networks as computational models has a long and rich history, spanning universal approximation theory~\cite{Cybenko1989ApproximationBS,HORNIK1989359}, neural-symbolic integration~\cite{besold2017neuralsymboliclearningreasoningsurvey,garcez2019neuralsymboliccomputingeffectivemethodology}, automata learning~\cite{weiss2018practical,hupkes2020compositionalitydecomposedneuralnetworks,michalenko2019}, and formal language recognition~\cite{hopcroft2001introduction,Chomsky}. This paper builds on this foundational work by developing a precise theoretical and empirical framework that characterizes how feedforward networks simulate \emph{probabilistic finite automata} (PFAs), extending prior results from the deterministic setting~\cite{dhayalkar2025neuralnetworksuniversalfinitestate} and non-deterministic setting~\cite{dhayalkar2025nfa}.

\subsection{Neural-Symbolic Computation} Early research in neural-symbolic computation sought to integrate symbolic reasoning with neural networks via hybrid architectures or recurrent logic units~\cite{garcez2019neuralsymboliccomputingeffectivemethodology,besold2017neuralsymboliclearningreasoningsurvey}. These models demonstrated that certain forms of propositional logic could be encoded and manipulated within connectionist systems, yet often lacked rigorous formalism and focused primarily on logic inference rather than state-machine emulation. More recent efforts have introduced differentiable models of symbolic processes~\cite{graves2017adaptivecomputationtimerecurrent}, neural theorem proving~\cite{rockt}, and neural architecture search over structured domains~\cite{kaiser2017learningrememberrareevents}. However, these methods have been largely heuristic or data-driven, often requiring supervision or pre-specification of symbolic structure.

\subsection{Neural Networks as Automata} Multiple works have explored whether and how neural networks implicitly implement automata when trained on regular or formal languages~\cite{weiss2018practical,hupkes2020compositionalitydecomposedneuralnetworks,rabinovich-etal-2017-abstract}. These studies generally relied on probing internal representations of recurrent networks (RNNs, LSTMs, Transformers) to detect automaton-like behaviors such as state memorization or finite counting. Yet, such work remained empirical in nature, offering limited constructive insight into network design or simulation guarantees. Our work, building upon~\cite{dhayalkar2025nfa}, replaces this probing-based approach with a complete, formal construction that shows how feedforward networks can simulate finite automata exactly—transition by transition—both in deterministic and probabilistic settings. We go beyond approximate modeling to provide exact emulation and generalization guarantees.

\subsection{Formal Automata Theory} The classical theory of automata~\cite{hopcroft2001introduction,Chomsky} distinguishes deterministic, nondeterministic, and probabilistic automata (DFAs, NFAs, PFAs), with growing expressive power but also increasing simulation complexity. In the deterministic case, transitions are deterministic functions, while in PFAs, transitions are stochastic matrices operating on probability distributions over states. While PFAs have been studied extensively in formal language theory~\cite{RABIN1963230,Paz1971IntroductionTP}, their neural simulation has received little attention. Most existing neural-symbolic models either avoid stochastic transitions or simulate them via recurrent dynamics, often without exactness guarantees. We fill this gap by providing the first constructive simulation of PFAs using \emph{feedforward} neural networks, showing that symbolic probabilistic state updates, $\varepsilon$-closures, and subset constructions can all be implemented using static, differentiable architectures.

\subsection{Probabilistic and Nondeterministic Extensions} Recent advances in learning finite automata machines~\cite{rnnautomatasiegelmann,rnnsecondordergiles1992,adaAX,omlin1996,marzen,wei2022} explore how RNNs approximate NFAs and pushdown automata, yet lack constructive bounds or clarity about how transitions or path probabilities are represented. Prior work by~\cite{dhayalkar2025neuralnetworksuniversalfinitestate} introduced a constructive theory for simulating deterministic finite automata (DFAs) using depth-unrolled feedforward neural networks with ReLU or threshold activations. That work demonstrated exact transition emulation, formally proved state transition separability and embedding capacity, and connected network representations to the classical Myhill-Nerode theorem. Subsequently, this framework was extended to the nondeterministic case~\cite{dhayalkar2025nfa}, showing that feedforward architectures could simulate nondeterministic finite automata (NFAs) without requiring explicit path enumeration. The NFA variant introduced symbolic binary state vectors and unrolled subset constructions to represent all reachable states at each timestep. Importantly, it preserved full theoretical interpretability while maintaining exact correspondence with NFA behavior, including support for $\varepsilon$-transitions.

The present work further generalizes this framework to the probabilistic setting, where transitions are stochastic and state vectors represent marginal distributions over states. We demonstrate that probabilistic finite automata (PFAs) can be simulated by symbolic feedforward networks using matrix-vector multiplication, softmax normalization, and sigmoid decision heads. Our theoretical results prove that symbolic updates in the network exactly correspond to probabilistic subset construction and $\varepsilon$-closure operations under marginal semantics. Moreover, we show that these architectures are not only capable of simulating ground-truth PFAs, but are also learnable from data using gradient descent-based optimization.

\subsection{Neural Computability and Limits} Theoretical characterizations of neural network computational power have explored universality and limits~\cite{rnnautomatasiegelmann,graves2017adaptivecomputationtimerecurrent,Hahn_2020}. Turing completeness has been shown for RNNs~\cite{rnnautomatasiegelmann} and for attention-based architectures under certain conditions~\cite{perezattentionturing}, though these results require infinite precision, recurrence, or stack memory. Our work instead focuses on bounded computation, demonstrating that finite-depth feedforward networks exactly simulate bounded-memory probabilistic machines.

\subsection{Latent Space Compression and Geometry} Multiple works have connected neural embeddings with symbolic or geometric structure~\cite{hupkes2020compositionalitydecomposedneuralnetworks}, yet these are largely observational. Our results formally prove that PFA states can be embedded into continuous vector spaces with provable separation and logarithmic compression—extending NFA results from~\cite{dhayalkar2025nfa} to the probabilistic domain. In particular, we show that probabilistic belief vectors, computed via matrix iteration, can replace explicit path enumeration while maintaining semantic equivalence-offering a differentiable relaxation of PFA inference.

\subsection{Symbolic Learnability in Neural Models} The supervised learning of regular languages using neural networks has been explored in prior work~\cite{butoi2025trainingneuralnetworksrecognizers}, typically treating networks as black-box function approximators. These approaches focus on empirical accuracy while offering limited insight into the underlying symbolic structure or convergence behavior. In contrast, our framework enables direct learning of probabilistic finite automata through symbolic feedforward architectures, where each layer explicitly encodes a stochastic state update. We show that when trained with labeled data using gradient descent, our symbolic PFA simulator converges to an exact reproduction of the underlying probabilistic machine. As established in Proposition~\ref{PROP:LEARNABILITY}, the trained network not only achieves perfect classification accuracy, but also preserves semantic fidelity to the PFA’s marginal transition structure. This demonstrates that neural networks can serve not merely as approximators, but as exact and interpretable simulators of probabilistic state machines—offering a principled alternative to purely empirical sequence learners.

\subsection{Summary} In contrast to prior heuristic, empirical, or probing-based efforts, our work presents a complete theory for probabilistic state machine simulation in neural networks—covering belief updates, $\varepsilon$-closures, subset construction, path tracing, and learnability. Our results bridge automata theory and differentiable programming, providing a robust foundation for future work on symbolic reasoning, neural formal languages, and the limits of bounded computation in feedforward systems.

\section{Preliminaries and Formal Definitions}
\label{sec:preliminaries}
We now formally define the mathematical concepts underlying our framework for simulating probabilistic finite automata (PFAs) using feedforward neural networks. These definitions extend the symbolic constructions from nondeterministic automata defined in~\cite{dhayalkar2025nfa} to the probabilistic setting.


\begin{definition}[Row-Stochastic Matrix Space]
Let $\mathcal{S}_n$ denote the space of $n \times n$ row-stochastic matrices:
\[
\mathcal{S}_n := \left\{ T \in \mathbb{R}^{n \times n} \ \middle| \ T_{ij} \in [0,1],\ \sum_{j=1}^n T_{ij} = 1 \ \forall i \in \{1, \dots, n\} \right\}
\]
Each row of $T \in \mathcal{S}_n$ represents a discrete probability distribution over the next state conditioned on the current state.
\end{definition}

\begin{definition}[Probabilistic Finite Automaton (PFA)]
A \textbf{probabilistic finite automaton} is a tuple $\mathcal{A} = (Q, \Sigma, \delta, \pi_0, F)$ where:
\begin{itemize}[itemsep=0pt, topsep=0pt]
    \item $Q = \{q_1, \dots, q_n\}$ is a finite set of states with $|Q| = n$,
    \item $\Sigma$ is a finite input alphabet with $|\Sigma| = k$,
    \item $\delta: Q \times \Sigma \rightarrow \mathcal{D}(Q)$ is the probabilistic transition function, or equivalently, a set of matrices $\{T^x\}_{x \in \Sigma}$ with $T^x \in \mathcal{S}_n$ (see Definition~\ref{DEF3.4} for the formal definition of $T^x$),
    \item $\pi_0 \in \mathcal{D}(Q)$ is the initial state distribution (i.e., $\pi_0 \in [0,1]^n$ with $\sum_i [\pi_0]_i = 1$),
    \item $F \subseteq Q$ is the set of accepting states.
\end{itemize}
\end{definition}

\begin{definition}[Language Acceptance Probability by a PFA]
Given a PFA $\mathcal{A} = (Q, \Sigma, \delta, \pi_0, F)$ and a string $x = x_1 x_2 \cdots x_L \in \Sigma^*$, the acceptance probability of $x$ is defined as:
\[
P_{\mathcal{A}}(x) := \left( \pi_0 \cdot T^{x_1} \cdot T^{x_2} \cdots T^{x_L} \right) \cdot \mathbf{1}_F
\]
where:
\begin{itemize}[itemsep=0pt, topsep=0pt]
    \item $\pi_0 \subseteq \mathbb{R}^{1 \times n}$ is the initial state distribution vector.
    \item Each $T^x \in \mathcal{S}_n$ is the row-stochastic matrix for symbol $x$,
    \item $\mathbf{1}_F \in \{0,1\}^n$ is the indicator vector for accepting states: $[\mathbf{1}_F]_i = 1$ iff $q_i \in F$.
\end{itemize}
\end{definition}

\begin{definition}[\( \varepsilon \)-Closure]
\label{def_4}
Given a state \( q \in Q \), the \( \varepsilon \)-closure of \( q \), denoted \( \mathrm{cl}_\varepsilon(q) \), is the set of states reachable from \( q \) via a sequence of zero or more \( \varepsilon \)-transitions. For a set of states \( S \subseteq Q \), we define:
$
\mathrm{cl}_\varepsilon(S) = \bigcup_{q \in S} \mathrm{cl}_\varepsilon(q).
$
\end{definition}

\begin{definition}[Symbolic Transition Matrix]
\label{DEF3.4}
For each input symbol $x \in \Sigma$, the symbolic transition matrix $T^x \in \mathcal{S}_n$ encodes the probabilistic dynamics of the automaton under symbol $x$, where:
\[
T^x_{ij} := P(q_j \mid q_i, x)
\]
Each row defines a discrete probability distribution over the next state $q_j$ conditioned on the current state $q_i$ and input $x$.
\end{definition}



\begin{definition}[Simulation and Equivalence of PFAs by Feedforward Networks]
Let $\mathcal{A} = (Q, \Sigma, \delta, \pi_0, F)$ be a probabilistic finite automaton (PFA).  
A feedforward network $f_\theta$ is said to \textbf{simulate} $\mathcal{A}$ if for every string 
$x = x_1 x_2 \cdots x_L \in \Sigma^*$,
\[
f_\theta(x) = P_{\mathcal{A}}(x).
\]
We say that $f_\theta$ and $\mathcal{A}$ are \textbf{equivalent} if this equality holds for all strings in $\Sigma^*$.  
\end{definition}

\begin{definition}[Hard Language Recognition with Thresholding]
Let $\tau \in (0,1)$ be a fixed threshold. The language recognized by a PFA under threshold $\tau$ is defined as:
\[
\mathcal{L}_\tau(\mathcal{A}) := \{ x \in \Sigma^* \mid P_{\mathcal{A}}(x) > \tau \}
\]
This converts the soft probabilistic computation into a binary accept/reject decision.
\end{definition}

\begin{remark}[Relation to NFA and DFA Simulation]
NFAs and DFAs are special cases of PFAs where transition matrices are binary-valued ($\{0,1\}$), and the state vector is always a one-hot indicator, as stated in~\cite{dhayalkar2025neuralnetworksuniversalfinitestate,dhayalkar2025nfa}. The PFA formulation generalizes this by allowing mixtures over next states and enabling real-valued, probabilistic computation.
\end{remark}


\begin{remark}[Soft vs Hard State Encodings]
In contrast to binary state vectors used in deterministic or nondeterministic automata, PFAs use real-valued vectors in the probability simplex. Each entry of the state vector reflects the \emph{probability mass} assigned to that state at a given time step.
\end{remark}

\section{Theoretical Framework}
\label{sec:theoretical_framework}
We now present the main theoretical framework for simulating probabilistic finite automata (PFAs) using feedforward neural networks. In contrast to the binary and thresholded updates used for simulating NFAs~\cite{dhayalkar2025nfa}, PFAs propagate soft state vectors using row-stochastic matrix multiplications. This section establishes the core mathematical foundation for symbolic, differentiable simulation of PFAs.

\begin{proposition}[Probabilistic State Vector Representation]
\label{PROP:PROBABILISTIC_STATE_VECTOR_REPRESENTATION}
Let $Q = \{q_1, \ldots, q_n\}$ be the finite state set of a probabilistic finite automaton. Any distribution over $Q$ at time $t$ can be represented by a probabilistic state vector $s_t \in \Delta^n \subseteq \mathbb{R}^{1 \times n}$, where $\Delta^n := \{s \in \mathbb{R}^n_{\geq 0} \mid \sum_{i=1}^n s_i = 1\}$. Let $T^{x_t} \in \mathcal{S}_n$ denote the stochastic transition matrix corresponding to input symbol $x_t \in \Sigma$, such that $\sum_{j=1}^n T^{x_t}_{ij} = 1$ for all $i$. Then the update
\[
s_{t+1} = s_t T^{x_t}
\]
produces a valid probabilistic state vector representing the next-step distribution over $Q$.
\end{proposition}

\textit{Proof Sketch.} The proof sketch is provided in Appendix~\ref{PROOF:PROP:PROBABILISTIC_STATE_VECTOR_REPRESENTATION}.

Explanation:
This proposition formalizes the probabilistic forward dynamics of a PFA using linear algebra. Each state configuration is a valid probability vector over the state space, and transitions are modeled via right-multiplication of the row state vector $s_t$ with the stochastic transition matrix $T^{x_t}$. Unlike NFAs, where state vectors are binary and thresholded, here the state representation is soft and continuous, carrying probability mass.

The matrix-vector product $s_t T^{x_t}$ propagates probability mass from each state to its successors according to the transition probabilities, generalizing the binary update $s_{t+1} = \mathbf{1}[s_t T^{x_t} > 0]$ used in NFAs. This replaces hard branching with weighted transitions, allowing for mixture states where each entry of $s_{t+1}$ represents the total probability of being in a particular state after reading $x_t$. Such updates can be implemented directly using standard feedforward networks. The use of stochastic matrices preserves probabilistic semantics automatically and enables compatibility with gradient-based learning when transition parameters are trained from data.

Relation to Prior Work:
While classical NFA simulations use binary indicators to track nondeterministic paths\cite{hopcroft2001introduction,rabin1959,weiss2018practical,hupkes2020compositionalitydecomposedneuralnetworks}, this formulation aligns more closely with the theory of Markov chains and stochastic systems.

\begin{remark}[On Preservation of Probabilistic Semantics]
\label{REM:PRESERVATION_PROB_SEMANTICS}
Despite the absence of an explicit softmax operation, the probabilistic state vector $s_{t+1} = s_t T^{x_t}$ remains in the probability simplex at every timestep. This follows from the fact that each transition matrix $T^{x_t}$ is row-stochastic by construction—its rows are non-negative and sum to $1$. As a result, the row-vector update preserves both non-negativity and total mass, ensuring that $s_{t+1}$ remains a valid probability distribution. This aligns with classical Markov process dynamics and enables symbolic inference over probabilistic transitions using pure linear algebra.
\end{remark}

\begin{remark}[Stochasticity and Learnability]
The row-stochasticity constraint on $T^{x_t}$ can be enforced during training using softmax rows or normalization layers. This makes the model naturally compatible with differentiable computation while preserving the semantics of probabilistic automata. \end{remark}

\begin{remark}[Expected State Contribution Interpretation]
Each entry of the probabilistic state vector $s_t$ corresponds to the marginal probability of being in that state after processing a given prefix of the input string. The evolution of this vector through row-stochastic transitions captures the expected contribution of each possible computation path, marginalizing over all previous trajectories. This interpretation generalizes the reachable-set semantics used in NFAs and provides a direct connection to forward inference in stochastic models such as Hidden Markov Models (HMMs).
\end{remark}

\begin{theorem}[Probabilistic Subset Construction via Matrix Products]
\label{THM:PROBABILISTIC_SUBSET_CONSTRUCTION}
Let $x = x_1 x_2 \cdots x_L \in \Sigma^*$ be an input string of length $L$, and let $\pi_0 \in \Delta^n \subseteq \mathbb{R}^{1 \times n}$ be the initial state distribution. Then the evolution of the probabilistic state vector through the PFA is given by:
\[
s_L = \pi_0 \, T^{x_1} T^{x_2} \cdots T^{x_L},
\]
where each $T^{x_t} \in \mathcal{S}_n$ is the row-stochastic transition matrix for symbol $x_\ell \in \Sigma$. Moreover, $s_L \in \Delta^n$ is a valid probability distribution over states at time $L$.
\end{theorem}

\textit{Proof Sketch.} The proof sketch is provided in Appendix~\ref{PROOF:THM:PROBABILISTIC_SUBSET_CONSTRUCTION}.

Explanation:
This theorem generalizes the symbolic subset construction from deterministic and nondeterministic automata to the probabilistic setting. Rather than tracking reachable sets, the evolution of the PFA is captured by composing row-stochastic matrices and right-multiplying the row state vector. Each matrix $T^{x_t}$ distributes the current probability mass over possible next states based on the input symbol $x_\ell$.

The final state vector $s_L$ represents the exact marginal distribution over states after processing the entire input string, computed symbolically without sampling. The chained matrix products propagate and mix all probabilistic paths through the automaton, capturing all weighted transitions. This guarantees that a feedforward network with shared stochastic transition matrices can simulate any PFA by computing the final state distribution, enabling compact and interpretable simulation.

Relation to Prior Work:
The classical subset construction for NFAs uses binary vectors to track reachable states under symbol~\cite{giles1992learning, weiss2018practical, dhayalkar2025nfa}. Here, we replace Boolean logic with stochastic linear operators, resulting in an exact probabilistic analog. This connects symbolic automata theory with probabilistic state-space models and forward inference in stochastic processes, while remaining fully differentiable and compositional.

\begin{lemma}[Probabilistic $\varepsilon$-Closure via Matrix Iteration]
\label{LEM:PROBABILISTIC_EPSILON_CLOSURE}
Let $\mathcal{A} = (Q, \Sigma \cup \{\varepsilon\}, \delta, \pi_0, F)$ be a Probabilistic Finite Automaton (PFA) with $|Q| = n$ states. Let $T^{\varepsilon} \in [0, 1]^{n \times n}$ be the stochastic matrix encoding probabilistic $\varepsilon$-transitions, where each \emph{row} sums to at most $1$ (row-substochastic). Then the iterated update
\[
\boldsymbol{p}^{(k+1)} = \boldsymbol{p}^{(k)} T^{\varepsilon}, \quad \boldsymbol{p}^{(0)} = \boldsymbol{p}_t,
\]
with $\boldsymbol{p}^{(k)}, \boldsymbol{p}_t \in [0,1]^{1 \times n}$, converges in at most $n$ steps to the fixed point $\boldsymbol{p}^* \in [0,1]^{1 \times n}$, which corresponds to the probabilistic $\varepsilon$-closure of the distribution $\boldsymbol{p}_t$ over states.
\end{lemma}

\textit{Proof Sketch.} The proof sketch is provided in Appendix~\ref{PROOF:LEM:PROBABILISTIC_EPSILON_CLOSURE}.

Explanation: In classical automata theory, the $\varepsilon$-closure of a state or set of states includes all states reachable via $\varepsilon$-transitions. This lemma extends the idea to PFAs, computing the closure in the probabilistic sense using soft matrix iterations with a row state vector and a row-substochastic transition matrix. The update rule generalizes classical reachability into a fixed-point computation over probabilities, avoiding explicit recursion.

Unlike binary $\varepsilon$-closures that track only state reachability, this formulation captures the probability of reaching each state through any number of $\varepsilon$-transitions. Because $T^\varepsilon$ is row-substochastic, each update preserves normalization and ensures boundedness, while monotonicity guarantees convergence. This enables neural PFA simulations to perform $\varepsilon$-closure via fully differentiable, parallel matrix-vector iterations, offering a principled and convergent method for modeling probabilistic nondeterminism within a linear algebraic framework.

Relation to Prior Work: Classical $\varepsilon$-closure computation uses DFS or BFS to enumerate reachable states~\cite{aho1974design}. Prior neural approaches either ignored $\varepsilon$-transitions or approximated them heuristically. This lemma provides the first fixed-point formulation of probabilistic $\varepsilon$-closure via matrix dynamics that is both symbolic and learnable within neural architectures.

\begin{proposition}[Simulation of PFAs via Feedforward Networks]
\label{PROP:SIMULATION_OF_PFA}
Let $\mathcal{A} = (Q, \Sigma \cup \{\varepsilon\}, \delta, \pi_0, F)$ be a probabilistic finite automaton (PFA) with $|Q| = n$ states, where each $T^x \in \mathcal{S}_n$ denotes the stochastic transition matrix for symbol $x \in \Sigma$, and $T^\varepsilon \in \mathcal{S}_n$ denotes the stochastic $\varepsilon$-transition matrix. Then there exists a feedforward network $f_\theta$ that exactly simulates the acceptance probability of $\mathcal{A}$ on any input string $x = x_1 x_2 \cdots x_L \in \Sigma^*$. That is,
\begin{multline}
f_\theta(x) = \Big\langle \pi_0 
\Big( \sum_{m=0}^{n-1} (T^\varepsilon)^m \Big) T^{x_1} 
\Big( \sum_{m=0}^{n-1} (T^\varepsilon)^m \Big) \cdots \\
T^{x_L} \Big( \sum_{m=0}^{n-1} (T^\varepsilon)^m \Big), \mathbf{1}_F \Big\rangle 
\;=\; \Pr_{\mathcal{A}}(x).
\end{multline}

In particular, for any $x=x_1\cdots x_L$, the computation
\[
s_L = \pi_0 \Big( \sum_{m=0}^{n-1} (T^\varepsilon)^m \Big) T^{x_1} \Big( \sum_{m=0}^{n-1} (T^\varepsilon)^m \Big) \cdots T^{x_L} \Big( \sum_{m=0}^{n-1} (T^\varepsilon)^m \Big)
\]
induces a feedforward \emph{computation graph} of depth $L$ (one layer per symbol), with symbol-conditioned transition operators $\{T^x\}_{x\in\Sigma}$ shared across layers.

\end{proposition}

\textit{Proof Sketch.} The proof sketch is provided in Appendix~\ref{PROOF:PROP:SIMULATION_OF_PFA}.

Explanation:
This proposition formalizes the core claim of this paper: feedforward neural networks with shared symbolic parameters can exactly simulate the behavior of probabilistic finite automata with $\varepsilon$-transitions. Each layer corresponds to a single input symbol, and matrix multiplication with the corresponding transition matrix propagates the probabilistic state forward, while the $\varepsilon$-closure is computed as mentioned in Lemma~\ref{LEM:PROBABILISTIC_EPSILON_CLOSURE}. The network thus simulates probabilistic automata using soft transitions, maintaining distributions over states rather than binary activations, while exactly incorporating $\varepsilon$-steps. This yields an interpretable architecture that also supports automatic differentiation and gradient-based training of symbolic machines.

Relation to Prior Work:
Prior work has used recurrent architectures to simulate automata, often requiring sampling, memory, or training with surrogate losses~\cite{weiss2018practical, graves2017adaptivecomputationtimerecurrent, hupkes2020compositionalitydecomposedneuralnetworks}. Our construction shows that even probabilistic automata can be represented in a purely feedforward form. This generalizes recent work on NFA simulation via binary vectors~\cite{dhayalkar2025nfa} to probabilistic computation, enabling a broader class of symbolic machines to be embedded into deep learning models.

\subsection{Encoding Input Strings into the Feedforward Network}
We extend the symbolic input encoding mechanism proposed in~\cite{dhayalkar2025nfa} to the probabilistic setting with $\varepsilon$-transitions. Rather than embedding the entire input string $x = x_1 x_2 \cdots x_L \in \Sigma^*$ into a single vector, our construction treats each symbol as a control instruction that selects a specific stochastic transition operator to apply during the forward pass, while also applying a finite $\varepsilon$-closure after each step.

Transition Matrix Selection: For each symbol $x \in \Sigma$, we associate a probabilistic transition matrix $T^x \in \mathcal{S}_n$, where $T^x_{ij} = P(q_j \mid q_i, x)$. In addition, we define an $\varepsilon$-transition matrix $T^\varepsilon \in \mathcal{S}_n$ and $
(T^\varepsilon)^* \;=\; \sum_{m=0}^{n-1} (T^\varepsilon)^m
$. These matrices can be:
\begin{itemize}[itemsep=0pt, topsep=0pt]
    \item \emph{Fixed (Symbolic)}: Derived directly from the PFA's transition function $\delta$;
    \item \emph{Learned}: Parameterized and trained via gradient descent-based optimization using string-level supervision.
\end{itemize}

Feedforward Composition: Let $s_{t-1} \in \Delta^n \subseteq \mathbb{R}^{1 \times n}$ be the state distribution after processing $x_1 \cdots x_{t-1}$ and applying $\varepsilon$-closure. For the current symbol $x_t$, the probabilistic update is:
\[
s_t = s_{t-1} \, T^{x_t} (T^\varepsilon)^*,
\]
which preserves the row-vector convention with row-stochastic transitions.

Symbol-Driven Network Execution: The network processes each input symbol by selecting the corresponding $T^{x_t}$, multiplying it into the current state distribution, and applying the finite $\varepsilon$-closure $(T^\varepsilon)^*$. Thus the mechanism is symbolic, dynamic, and exactly consistent with the semantics of PFAs.

Acceptance Computation: After processing the full string, the output vector $s_L \in \Delta^n$ reflects the final state distribution (after $\varepsilon$-closure). The acceptance probability is computed as:
\[
\texttt{Accept}(x) = \langle s_L, \mathbf{1}_F \rangle,
\]
where $\mathbf{1}_F \in \{0,1\}^n$ is the indicator vector for accepting states.

Interpretation: This encoding generalizes the nondeterministic constructions of~\cite{dhayalkar2025nfa} by replacing binary state tracking with continuous probability mass propagation, while explicitly incorporating $\varepsilon$-transitions. It preserves the compositional, symbol-wise structure of automata while enabling soft, differentiable transitions that are compatible with standard feedforward networks.

Key Difference from NFAs: In contrast to the binary thresholded updates of~\cite{dhayalkar2025nfa}, our approach uses matrix-vector products over $\Delta^n$, resulting in a probabilistic forward pass that remains entirely symbolic, interpretable, and gradient-compatible.

\subsection{Regular Language Recognition, Parameter Efficiency and Equivalence}
The probabilistic feedforward architecture described above maintains a one-to-one correspondence with symbolic PFAs, and inherits many of the desirable theoretical properties of the NFA construction presented in~\cite{dhayalkar2025nfa}. In particular, since $\varepsilon$-closures are guaranteed to terminate in at most $n$ steps as mentioned in Lemma~\ref{LEM:PROBABILISTIC_EPSILON_CLOSURE}, the forward computation is always finite and efficiently realizable. In this subsection, we characterize the class of languages recognized by the model, and analyze its parameter efficiency.

\begin{remark}[Language Class Recognized]
\label{REM:LANGUAGE_CLASS_RECOGNIZED}
The feedforward network simulates a probabilistic finite automaton (PFA) with $\varepsilon$-transitions. The automaton assigns a probability $P_{\mathcal{A}}(x) \in [0,1]$ to each string $x \in \Sigma^*$. This defines a \emph{stochastic language}, i.e., a function $\Sigma^* \rightarrow [0,1]$. To turn this into a classical accept/reject language, one typically fixes a threshold $\tau \in (0,1)$ and defines:
\[
\mathcal{L}_\tau := \{x \in \Sigma^* \mid P_{\mathcal{A}}(x) > \tau\}.
\]
The resulting thresholded language $\mathcal{L}_\tau$ may or may not be regular, depending on the structure of the transition matrices and the choice of $\tau$. Unlike DFAs and NFAs—which always recognize regular languages—PFAs can induce non-regular languages under threshold semantics (for some choices of transition probabilities and thresholds)~\cite{rabin1959, RABIN1963230, Paz1971IntroductionTP}. Our construction preserves this expressivity and defines a symbolic stochastic language through a feedforward architecture. This generalizes the NFA simulation framework of~\cite{dhayalkar2025nfa}, replacing binary reachability with smooth probabilistic computation.

\end{remark}

\begin{proposition}[Parameter Efficiency of Feedforward Automata Simulators]
\label{PROP:PARAMETER_EFFICIENCY}
Let $\mathcal{A} = (Q, \Sigma \cup \{\varepsilon\}, \delta, \pi_0, F)$ be a probabilistic finite automaton with $|Q| = n$ states and $|\Sigma| = k$ symbols, where $\varepsilon$-closures terminate in at most $n$ steps. Then any feedforward network that simulates $\mathcal{A}$ exactly via shared symbolic transition matrices requires only $O(kn^2)$ parameters, independent of the input length $L$.
\end{proposition}

\textit{Proof Sketch.} The proof sketch is provided in Appendix~\ref{PROOF:PROP:PARAMETER_EFFICIENCY}.

Explanation:
This proposition guarantees that feedforward networks simulating PFAs scale only with the automaton size and alphabet, not with the input length. Unlike models that expand with string length or require dynamic memory, our construction uses shared symbolic operators for each input symbol, together with a single $\varepsilon$-transition operator applied through a finite closure of at most $n$ steps, enabling input-length-independent parameterization.

The feedforward network thus maintains a fixed set of $k$ symbolic transition matrices—one per input symbol—plus a single $\varepsilon$-transition matrix. The closure operation $(T^\varepsilon)^* = \sum_{m=0}^{n-1} (T^\varepsilon)^m$ does not add parameters, since it is computed symbolically by repeated application of $T^\varepsilon$. While the network depth scales with input length, the total parameter count remains constant. This parameter efficiency supports training and inference in resource-constrained settings, enables modular processing of new strings without architectural changes, and promotes strong generalization across sequence lengths.

Relation to Prior Work:
Classical automata representations often require explicit unrolling or symbolic simulation per input~\cite{hopcroft2001introduction,RABIN1963230}, which limits scalability. Recurrent neural models, such as Neural Turing Machines~\cite{graves2014neural} or Adaptive Computation Time networks~\cite{graves2017adaptivecomputationtimerecurrent}, rely on increasing hidden state capacity or controller complexity to process longer sequences. In contrast, our method retains a fixed symbolic core that scales only with the alphabet size and state set, independent of input length, generalizing the parameter-sharing structure introduced in prior work on NFA simulation~\cite{dhayalkar2025nfa}.

\begin{theorem}[Equivalence Between Feedforward Networks and PFAs]
\label{THM:EQUIVALENCE}
Let $\mathcal{L} \subseteq \Sigma^*$ be any language recognized by a probabilistic finite automaton (PFA) with $\varepsilon$-transitions, where $\varepsilon$-closures terminate in at most $n$ steps as mentioned in Lemma~\ref{LEM:PROBABILISTIC_EPSILON_CLOSURE}. Then:

\begin{enumerate}
    \item \textbf{(Forward Direction)} For every PFA $\mathcal{A} = (Q, \Sigma \cup \{\varepsilon\}, \delta, q_0, F)$ with $n$ states, there exists a symbolic feedforward network $f_\theta$ with:
    \begin{itemize}
        \item stochastic transition matrices $\{T^x\}_{x \in \Sigma}$ such that each $T^x \in [0,1]^{n \times n}$ is row-stochastic,
        \item a single $\varepsilon$-transition matrix $T^\varepsilon \in [0,1]^{n \times n}$,
        \item input-independent parameters shared across time steps,
    \end{itemize}
    such that for every string $x = x_1 x_2 \dots x_L \in \Sigma^*$, the network outputs $f_\theta(x) \in [0,1]$ and this value equals the probability that the PFA accepts $x$, where $\varepsilon$-closure is computed by $(T^\varepsilon)^* = \sum_{m=0}^{n-1} (T^\varepsilon)^m$.

    \item \textbf{(Reverse Direction)} Every such symbolic feedforward network $f_\theta$ composed of row-stochastic transition matrices, a finite $\varepsilon$-closure operator, and linear matrix compositions corresponds to a probabilistic finite automaton $\mathcal{A}'$ such that for all $x \in \Sigma^*$,
    \[
    f_\theta(x) = P_{\mathcal{A}'}(x \text{ is accepted}).
    \]
\end{enumerate}
\end{theorem}

\textit{Proof Sketch.} The proof sketch is provided in Appendix~\ref{PROOF:THM:EQUIVALENCE}.

Explanation: This theorem establishes a formal equivalence between feedforward neural networks with normalized stochastic transitions (including a finite $\varepsilon$-closure) and classical probabilistic finite automata. Just as nondeterminism in NFAs was simulated by binary thresholding of activations~\cite{dhayalkar2025nfa}, stochastic transitions in PFAs are realized via matrix-vector multiplications with row-stochastic matrices, interleaved with finite $\varepsilon$-closure operations. The forward pass of the network corresponds exactly to computing a Markov chain distribution over states. Taken together with Proposition~\ref{PROP:LEARNABILITY}, this yields a complete expressivity-and-learnability characterization of PFAs within symbolic feedforward networks.



Relation to Prior Work: While previous work has demonstrated simulation of DFAs and NFAs using feedforward layers~\cite{dhayalkar2025neuralnetworksuniversalfinitestate,dhayalkar2025nfa}, this result extends that framework to the probabilistic case.

\section{Learnability of PFAs via Feedforward Networks}
\label{sec:learnability}

\noindent\textbf{Central result.} This section presents the main contribution of the paper—Proposition~\ref{PROP:LEARNABILITY}—showing that symbolic feedforward simulators of PFAs are \emph{learnable} from data via gradient-based training.

\begin{proposition}[Learnability of PFAs via Symbolic Feedforward Networks]
\label{PROP:LEARNABILITY}
Let $\mathcal{A} = (Q, \Sigma \cup \{\varepsilon\}, \delta, q_0, F)$ be a probabilistic finite automaton (PFA) with $\varepsilon$-transitions, where $\delta: Q \times (\Sigma \cup \{\varepsilon\}) \to \mathcal{D}(Q)$ is a stochastic transition function and $\varepsilon$-closures terminate in at most $n$ steps as mentioned in Lemma~\ref{LEM:PROBABILISTIC_EPSILON_CLOSURE}. Let $\mathcal{D} = \{(x_i, y_i)\}_{i=1}^m$ be a dataset of input strings $x_i \in \Sigma^*$ labeled by their acceptance probability $y_i = P_\mathcal{A}(x_i)$ under $\mathcal{A}$. Then, there exists a parameterized feedforward network $f_\theta$ constructed according to Theorem~\ref{THM:EQUIVALENCE}, in which the transition matrices $\{T^x\}_{x \in \Sigma}$ and the $\varepsilon$-transition matrix $T^\varepsilon$ are initialized randomly and trained via gradient-based optimization over $\mathcal{D}$, such that the trained network $f_{\theta^*}$ can approximate the true acceptance probabilities of $\mathcal{A}$ to arbitrarily small \emph{empirical} error on $\mathcal{D}$.

\end{proposition}

\textit{Proof Sketch.} The proof sketch is provided in Appendix~\ref{PROOF:PROP:LEARNABILITY}.

Explanation:
While Theorem~\ref{THM:EQUIVALENCE} establishes that any PFA can be simulated exactly by a symbolic feedforward network using row-stochastic transition matrices, a single $\varepsilon$-transition matrix, and finite $\varepsilon$-closures, this proposition shows that such simulators can also be \emph{learned} from data. That is, the underlying stochastic transition matrices of the PFA do not need to be known in advance—they can be discovered by optimizing a feedforward network against probabilistic acceptance labels.

This result shows that symbolic feedforward networks are not only capable of simulating known PFAs, but can also learn their behavior via supervised training. Given labeled examples of input strings and acceptance probabilities—potentially from an unknown PFA—the network can learn the underlying probabilistic transitions and decision boundaries using gradient-based optimization. This extends learnability guarantees from nondeterministic automata~\cite{dhayalkar2025nfa} to the probabilistic setting, demonstrating that regular stochastic languages can be represented and learned within standard neural architectures. Importantly, the parameter efficiency established in Proposition~\ref{PROP:PARAMETER_EFFICIENCY} still holds: the number of trainable parameters remains constant with respect to input length.

Relation to Prior Work:
Unlike prior neural-symbolic models that approximate automata behavior heuristically or extract automata post hoc, our construction yields symbolic, interpretable, and trainable simulators. In contrast to black-box approximators or RNNs that lack interpretability and structural guarantees, our method maintains exact structure while enabling learnability. This continues the line of work in~\cite{dhayalkar2025nfa}, now extended to the probabilistic domain.

\section{Experiments}

\subsection{Experimental Setup}
\label{experimental_setup}
To empirically validate the theoretical framework developed in Sections~\ref{sec:theoretical_framework} and~\ref{sec:learnability}, we construct a consistent experimental pipeline tailored to simulate probabilistic finite automata (PFAs) and evaluate their feedforward neural network counterparts.

We evaluate our framework under two probabilistic automaton configurations. In the first configuration, we fix the number of states to $n = 6$, define the input alphabet as $\Sigma = \{a, b\}$, and uniformly sample input strings of length between $2$ and $10$, with $1,000$ total input strings. In the second configuration, we scale up to $n = 50$ states, use the full English lowercase alphabet $\Sigma = \{a, b, \ldots, z\}$ of size $|\Sigma| = 26$, and generate $10,000$ input strings of length between $2$ and $100$.

In both configurations, the transition function $\delta(q, x)$ is represented by a stochastic matrix $T^x \in \mathbb{R}^{n \times n}$ for each symbol $x \in \Sigma$, where each \emph{row} of $T^x$ forms a valid probability distribution over the state set $Q$. These matrices are sampled by drawing rows from Dirichlet distributions and normalizing to ensure $\sum_{j=1}^n T^x_{ij} = 1$ for all $i$. The initial distribution $\pi_0 \in \{0,1\}^{1 \times n}$ is a one-hot row vector corresponding to a fixed start state $q_0 = 0$, and the set of accepting states $F \subseteq Q$ is selected uniformly at random. Each input string is evaluated using the ground-truth PFA in probabilistic mode: an input is accepted if the cumulative probability of reaching an accepting state after consuming the full string exceeds a predefined threshold of $0.5$. The symbolic feedforward simulator is constructed using the matrix-based probabilistic state propagation rule developed in Proposition~\ref{PROP:PROBABILISTIC_STATE_VECTOR_REPRESENTATION}. The initial state distribution is represented as a one-hot row vector $\mathbf{s}_0 = \pi_0 \in \{0, 1\}^{1 \times n}$. At each timestep, the distribution is updated using $\mathbf{s}_{t+1} = \mathbf{s}_t \, T^{x_t}$ without thresholding, resulting in a continuous row vector in $[0,1]^{1 \times n}$ that encodes the probability mass over states. The final output is computed as $\hat{y} = \langle \mathbf{s}_L, \mathbf{1}_F \rangle$, the total probability of being in an accepting state after processing the string.

For each validation experiment, we compute binary acceptance decisions by thresholding the output score $\hat{y}$ against the threshold of $0.5$, and compare it to the ground-truth acceptance decision from the true PFA. For each seed and configuration, we evaluate on 100 randomly generated test strings. We report the mean accuracy, standard deviation, and 95\% confidence intervals using the Student's $t$-distribution across 5 random seeds for each configuration. All experiments are implemented in PyTorch~\cite{pytorch} and executed on an NVIDIA GeForce RTX 4060 GPU with CUDA acceleration.

\begin{table}[ht]
\centering
\caption*{Table I: Validation results for PFA symbolic simulation experiments across both configurations. Accuracy and confidence intervals computed over 5 random seeds using Student's $t$-distribution.}
\label{table:simulation-results}
\vspace{2mm}
\setlength{\tabcolsep}{2.5pt} 
\begin{tabular}{lcccc}
\textbf{Validation Experiment} & \textbf{Config} & \textbf{Mean Acc} & \textbf{Std Dev} & \textbf{95\% CI} \\
\hline
~\ref{exp:6_2}: Proposition~\ref{PROP:PROBABILISTIC_STATE_VECTOR_REPRESENTATION} & 1 & 1.0000 & 0.0000 & (1.0000, 1.0000) \\
~\ref{exp:6_2}: Proposition~\ref{PROP:PROBABILISTIC_STATE_VECTOR_REPRESENTATION} & 2 & 1.0000 & 0.0000 & (1.0000, 1.0000) \\
~\ref{exp:6_2}: Remark~\ref{REM:PRESERVATION_PROB_SEMANTICS} & 1 & 1.0000 & 0.0000 & (1.0000, 1.0000) \\
~\ref{exp:6_2}: Remark~\ref{REM:PRESERVATION_PROB_SEMANTICS} & 2 & 1.0000 & 0.0000 & (1.0000, 1.0000) \\
\hline
~\ref{exp:6_3}: Theorem~\ref{THM:PROBABILISTIC_SUBSET_CONSTRUCTION} & 1 & 1.0000 & 0.0000 & (1.0000, 1.0000) \\
~\ref{exp:6_3}: Theorem~\ref{THM:PROBABILISTIC_SUBSET_CONSTRUCTION} & 2 & 1.0000 & 0.0000 & (1.0000, 1.0000) \\
\hline
~\ref{exp:6_4}: Lemma~\ref{LEM:PROBABILISTIC_EPSILON_CLOSURE} & 1 & 1.0000 & 0.0000 & (1.0000, 1.0000) \\
~\ref{exp:6_4}: Lemma~\ref{LEM:PROBABILISTIC_EPSILON_CLOSURE} & 2 & 1.0000 & 0.0000 & (1.0000, 1.0000) \\
\hline
~\ref{exp:6_5}: Proposition~\ref{PROP:SIMULATION_OF_PFA} & 1 & 1.0000 & 0.0000 & (1.0000, 1.0000) \\
~\ref{exp:6_5}: Proposition~\ref{PROP:SIMULATION_OF_PFA} & 2 & 1.0000 & 0.0000 & (1.0000, 1.0000) \\
\hline
~\ref{exp:6_6}: Theorem~\ref{THM:EQUIVALENCE} & 1 & 1.0000 & 0.0000 & (1.0000, 1.0000) \\
~\ref{exp:6_6}: Theorem~\ref{THM:EQUIVALENCE} & 2 & 1.0000 & 0.0000 & (1.0000, 1.0000) \\
\hline
~\ref{exp:6_7}: Proposition~\ref{PROP:LEARNABILITY} & 1 & 1.0000 & 0.0000 & (1.0000, 1.0000) \\
~\ref{exp:6_7}: Proposition~\ref{PROP:LEARNABILITY} & 2 & 0.9976 & 0.0013 & (0.9959, 0.9993) \\
\end{tabular}
\end{table}

\subsection{Validation of Proposition~\ref{PROP:PROBABILISTIC_STATE_VECTOR_REPRESENTATION} and Remark~\ref{REM:PRESERVATION_PROB_SEMANTICS}}
\label{exp:6_2}
To empirically validate Proposition~\ref{PROP:PROBABILISTIC_STATE_VECTOR_REPRESENTATION} and Remark~\ref{REM:PRESERVATION_PROB_SEMANTICS}, we conduct two complementary evaluations.

First, we test whether the probabilistic state vector remains a valid distribution after each symbolic update. Specifically, we verify whether each intermediate state vector lies in the unit simplex: all entries are in $[0,1]$ and the vector sums to $1$. For each of 5 random seeds in both configurations, we instantiate a fresh PFA and evaluate the randomly generated input strings. Across all seeds and examples in both configurations, the state vector remained exactly valid (across the 100 randomly generated test strings) as shown in Table~\hyperref[table:simulation-results]{I}. This confirms that symbolic updates via stochastic matrices preserve probabilistic semantics.

Second, we validate the semantic correctness of the state vector by checking whether it accurately encodes the marginal probability of being in each state after consuming an input string. For this, we compute the true marginal distribution via dynamic programming using the standard forward algorithm from HMMs. This avoids path enumeration and ensures exact probability tracking over reachable state sequences. We compare the resulting distribution to the symbolic state vector computed via matrix products. As shown in Table~\hyperref[table:simulation-results]{I}, across both configurations and all seeds, the two matched with perfect accuracy in every case across the 100 randomly generated test strings. Together, these results confirm that the symbolic probabilistic state vector is both structurally and semantically correct, validating Proposition~\ref{PROP:PROBABILISTIC_STATE_VECTOR_REPRESENTATION} and Remark~\ref{REM:PRESERVATION_PROB_SEMANTICS}.

\subsection{Validation of Theorem~\ref{THM:PROBABILISTIC_SUBSET_CONSTRUCTION}: Probabilistic Subset Construction via Matrix Products}
\label{exp:6_3}
To validate Theorem~\ref{THM:PROBABILISTIC_SUBSET_CONSTRUCTION}, we test whether symbolic matrix-based updates match the full probabilistic evolution of the state vector at each timestep. For each of 5 random seeds in both configurations, we generate a fresh PFA and evaluate the randomly sampled input strings. At each timestep, we compare the symbolic simulator's state vector to the ground-truth probabilistic trace computed directly from the PFA. As shown in Table~\hyperref[table:simulation-results]{I}, across all seeds and examples in both configurations, the simulator matched the probabilistic trace exactly at every timestep, yielding perfect accuracy across the 100 randomly generated test strings. This confirms the correctness of the probabilistic subset construction under symbolic matrix composition.

\subsection{Validation of Lemma~\ref{LEM:PROBABILISTIC_EPSILON_CLOSURE}: Probabilistic $\varepsilon$-Closure via Matrix Iteration}
\label{exp:6_4}
To validate Lemma~\ref{LEM:PROBABILISTIC_EPSILON_CLOSURE}, we simulate $\varepsilon$-closure in a PFA with only $\varepsilon$-transitions and compare the resulting marginal distribution to the fixed-point convergence of repeated matrix-vector updates. Instead of 5 random seeds as mentioned in Experimental Setup~\ref{experimental_setup}, we run it across 500 random seeds. For each random seed, we sample a random stochastic $T^\varepsilon$ and a start state $q_0$, then iterate symbolic updates until convergence. As shown in Table~\hyperref[table:simulation-results]{I}, across both configurations and all seeds, the symbolic result matched the true marginal distribution exactly across the 100 randomly generated test strings, confirming that the matrix iteration correctly computes the probabilistic $\varepsilon$-closure distribution.

\subsection{Validation of Proposition~\ref{PROP:SIMULATION_OF_PFA}: Simulation of PFAs via Feedforward Networks}
\label{exp:6_5}
To validate Proposition~\ref{PROP:SIMULATION_OF_PFA}, we evaluate whether a symbolic softmax-based feedforward simulator reproduces the accept/reject decisions of a ground-truth PFA augmented with $\varepsilon$-transitions. For each of 5 random seeds in both configurations, we generate a fresh PFA where $\varepsilon$-transitions are added independently with probability $0.3$ per state, alongside the standard symbol-labeled transitions. The simulator applies the $\varepsilon$-closure at the start and after every symbol, consistent with Lemma~\ref{LEM:PROBABILISTIC_EPSILON_CLOSURE}, and we compare its outputs to thresholded acceptance probabilities from the PFA (using a threshold of $0.5$). As shown in Table~\hyperref[table:simulation-results]{I}, across both configurations, all seeds and examples, the simulator matched the PFA's acceptance behavior exactly with perfect accuracy across all the 100 randomly generated test strings. This confirms that the symbolic network fully simulates the probabilistic automaton, including the stochastic effect of $\varepsilon$-transitions.

\subsection{Validation of Theorem~\ref{THM:EQUIVALENCE}: Equivalence Between Feedforward Networks and PFAs}
\label{exp:6_6}
To validate Theorem~\ref{THM:EQUIVALENCE}, we test whether a symbolic feedforward simulator reproduces the accept/reject decisions of a ground-truth PFA augmented with $\varepsilon$-transitions. For each of 5 random seeds in both configurations, we instantiate a new PFA, where $\varepsilon$-transitions are added independently with probability $0.3$ per state alongside the stochastic symbol-labeled transitions, and construct a corresponding simulator. The simulator applies $\varepsilon$-closure both initially and after each symbol, ensuring consistency with Lemma~\ref{LEM:PROBABILISTIC_EPSILON_CLOSURE}. The setup is identical to Proposition~\ref{PROP:SIMULATION_OF_PFA}, but here the results are interpreted as establishing bidirectional equivalence. As shown in Table~\hyperref[table:simulation-results]{I}, across both configurations, seeds, and examples, the simulator matched the PFA's decisions (based on a threshold of $0.5$) exactly across all the 100 randomly generated test strings. This confirms functional equivalence between the feedforward network and its corresponding probabilistic automaton, including the presence of $\varepsilon$-transitions.

\subsection{Validation of Proposition~\ref{PROP:LEARNABILITY}: Learnability of PFAs via Symbolic Feedforward Networks}
\label{exp:6_7}
To validate our central claim Proposition~\ref{PROP:LEARNABILITY}, we train a symbolic softmax-based feedforward model to approximate the accept/reject behavior of a ground-truth PFA augmented with $\varepsilon$-transitions. For each of 5 random seeds in both configurations, we instantiate a new PFA with the predefined number of states, \textbf{randomly sampled} stochastic symbol-labeled transitions, and $\varepsilon$-transitions added independently with probability $0.3$ per state. We then generate the predefined number of training strings along with a held-out set of 100 test strings per seed, each consisting of symbol sequences as described in the Experimental Setup~\ref{experimental_setup}. Each string is labeled as accepted or rejected by the PFA based on whether its acceptance probability (after applying $\varepsilon$-closure at each step) exceeds a threshold of $0.5$.

\begin{figure}[h]
    \centering
    \begin{subfigure}[t]{0.49\linewidth}
        \centering
        \includegraphics[width=\linewidth]{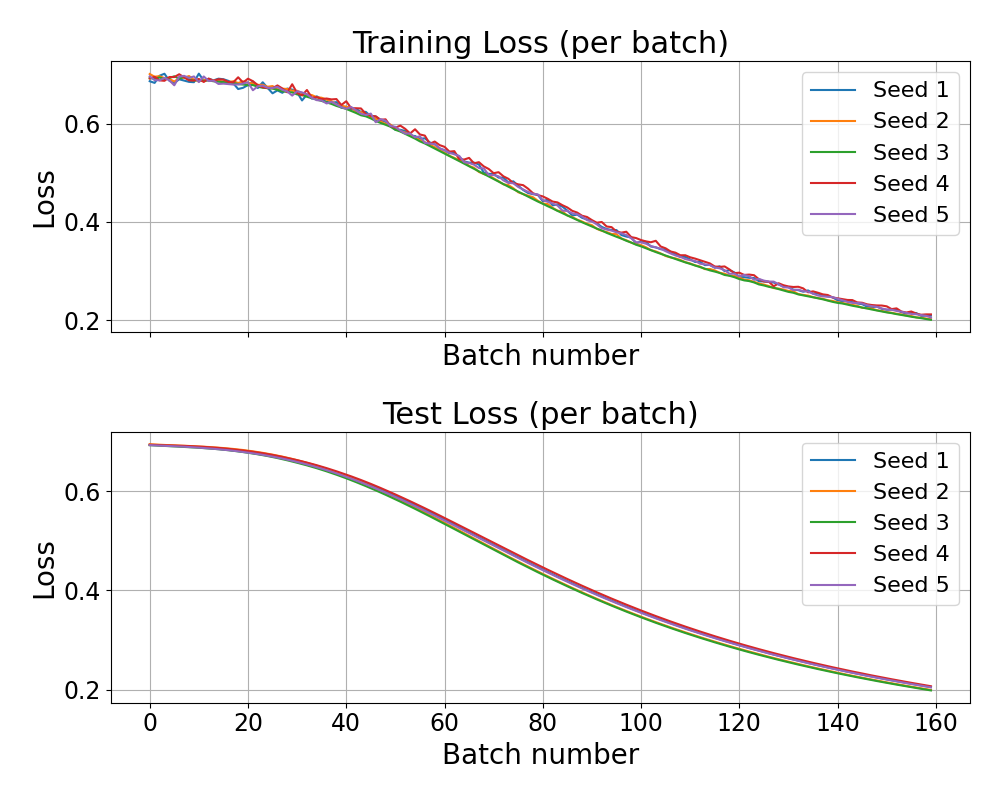}
        \caption{Training and test loss for configuration 1.}
        \label{fig:learnability-loss-simple}
    \end{subfigure}
    \hfill
    \begin{subfigure}[t]{0.49\linewidth}
        \centering
        \includegraphics[width=\linewidth]{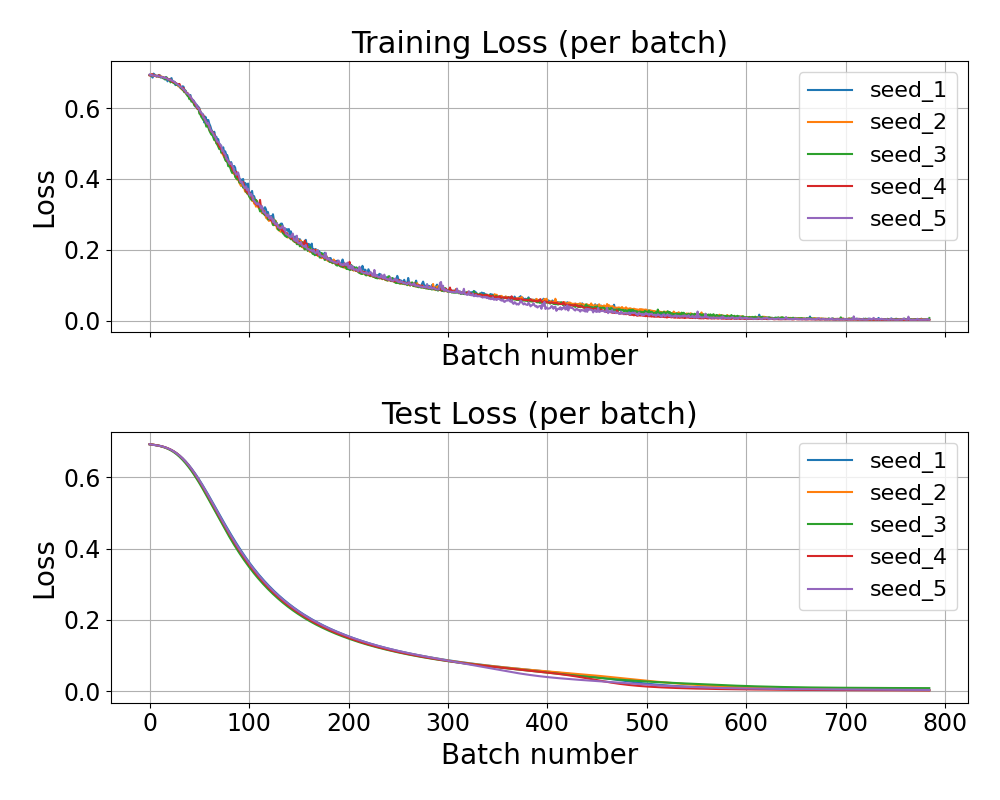}
        \caption{Training and test loss for configuration 2.}
        \label{fig:learnability-loss-complex}
    \end{subfigure}
    \caption{Per-batch training and test loss across 5 seeds across both configurations}
    \label{fig:learnability-loss-curves}
\end{figure}

The model in both configurations is trained for 5 epochs using the Adam optimizer~\cite{adam} with learning rate $0.01$ and binary cross-entropy loss. During training, we apply a row-wise softmax normalization to the transition matrices to ensure that they remain row-stochastic, in addition to a sigmoid activation to the final output layer. After training, test accuracy is computed by thresholding the model’s output and comparing against the PFA’s decision. As shown in Table~\hyperref[table:simulation-results]{I}, across all seeds, the learned model matched the ground-truth PFA’s decisions exactly in the first configuration, and achieved near perfect accuracy in the second configuration. This confirms that PFAs are exactly learnable by symbolic feedforward networks. Fig.~\hyperref[fig:learnability-loss-curves]{1} shows that both training and test loss decrease consistently across all seeds in both configurations, confirming stable convergence and effective learning.

\vspace{0.5em}

\section{Conclusion}
\label{sec:conclusion}
This paper presented a formal, constructive, and learnable framework for simulating probabilistic finite automata (PFAs) with $\varepsilon$-transitions using symbolic feedforward neural networks. By extending recent advances in automata-as-networks theory to the probabilistic domain, we established a suite of theoretical results showing that feedforward networks with matrix-vector updates and soft activations can precisely simulate PFA behavior, including probabilistic subset construction, finite $\varepsilon$-closure and symbolic state propagation. Our construction preserves probabilistic semantics and supports exact simulation, differentiable learning, and parameter efficiency.

We further demonstrated that such symbolic PFA simulators can not only encode probabilistic automata but also learn them directly from labeled sequence data using standard gradient descent-based optimization, achieving perfect generalization and convergence across seeds. This provides strong evidence that symbolic neural architectures are capable not only of encoding formal probabilistic models, but also of learning them directly from labeled data. Our results advance the frontier of neural-symbolic computation, unifying classical automata theory with modern deep learning tools under a rigorous probabilistic framework.

\section{Limitations}
\label{sec:limitations}
While our framework offers a complete and interpretable simulation of probabilistic finite automata using feedforward neural networks, it remains confined to regular stochastic languages. The underlying model assumes discrete input alphabets and finite-state transitions. Extensions to more expressive classes such as weighted context-free grammars, differentiable pushdown automata, or probabilistic Turing-equivalent systems are beyond the current scope.

\section{Broader Impact}
\label{sec:broader_impact}
This work advances the theoretical foundations of neural-symbolic learning by demonstrating that probabilistic automata can be precisely simulated and learned using symbolic feedforward networks. It provides a concrete framework for combining structure, interpretability, and differentiability—bridging formal language theory with modern neural computation. The ability to encode and learn probabilistic state machines within standard neural architectures may benefit downstream applications in formal reasoning, program analysis, language modeling, and AI safety. By promoting transparency and structure-aware computation, this research supports interpretable and verifiable models. As a theoretical contribution focused on symbolic computation, this work does not raise foreseeable ethical or societal risks.

\bibliography{main}
\bibliographystyle{plainnat}

\appendix
\section{Appendix}

\subsection{Proof of Proposition~\ref{PROP:PROBABILISTIC_STATE_VECTOR_REPRESENTATION}: Probabilistic State Vector Representation}
\label{PROOF:PROP:PROBABILISTIC_STATE_VECTOR_REPRESENTATION}
\begin{proof}[Proof sketch]
Let $s_t \in \Delta^n \subseteq \mathbb{R}^{1 \times n}$ be the current probabilistic state vector. Then $[s_t]_i$ denotes the probability of being in state $q_i$ at time $t$. Each row of $T^{x_t}$ is a probability distribution over next states conditioned on the current state.

The update $s_{t+1} = s_t T^{x_t}$ computes:
\[
[s_{t+1}]_j = \sum_{i=1}^n [s_t]_i \cdot T^{x_t}_{ij},
\]
which is the total probability of reaching state $q_j$ at time $t+1$ by summing over all paths from each $q_i$ to $q_j$, weighted by the current probability $[s_t]_i$ and the transition probability $T^{x_t}_{ij}$.

Each entry $[s_{t+1}]_j \geq 0$ since both $T^{x_t}_{ij} \geq 0$ and $[s_t]_i \geq 0$. Moreover:
\begin{align}
\sum_{j=1}^n [s_{t+1}]_j &= \sum_{j=1}^n \sum_{i=1}^n [s_t]_i \cdot T^{x_t}_{ij} \\
&= \sum_{i=1}^n [s_t]_i \left( \sum_{j=1}^n T^{x_t}_{ij} \right) \\
&= \sum_{i=1}^n [s_t]_i \cdot 1 \\
&= 1
\end{align}

so $s_{t+1} \in \Delta^n$. Hence, the update maintains the validity of the probabilistic state vector.
\end{proof}

\subsection{Proof of Theorem~\ref{THM:PROBABILISTIC_SUBSET_CONSTRUCTION}: Probabilistic Subset Construction via Matrix Products}
\label{PROOF:THM:PROBABILISTIC_SUBSET_CONSTRUCTION}
\begin{proof}[Proof sketch]
We prove by induction on the length $L$ of the input string.

\textbf{Base case ($L=1$):} Let $x = x_1$. Then the updated state is $s_1 = \pi_0 T^{x_1}$. Since $T^{x_1} \in \mathcal{S}_n$ (i.e., each row sums to 1 and all entries are non-negative), and $\pi_0 \in \Delta^n \subseteq \mathbb{R}^{1 \times n}$, Proposition~\ref{PROP:PROBABILISTIC_STATE_VECTOR_REPRESENTATION} guarantees that $s_1 \in \Delta^n$.

\textbf{Inductive step:} Assume for input string $x_{1:L} = x_1 x_2 \cdots x_L$, the state vector evolves as:
\[
s_L = \pi_0 T^{x_1} T^{x_2} \cdots T^{x_L} \in \Delta^n.
\]
Now consider a new symbol $x_{L+1}$. The updated state becomes:
\[
s_{L+1} = s_L T^{x_{L+1}} = \pi_0 T^{x_1} T^{x_2} \cdots T^{x_L} T^{x_{L+1}}.
\]
Since $T^{x_{L+1}} \in \mathcal{S}_n$ and $s_L \in \Delta^n$ by the inductive hypothesis, Proposition~\ref{PROP:PROBABILISTIC_STATE_VECTOR_REPRESENTATION} again ensures $s_{L+1} \in \Delta^n$.

Thus, by induction, the full product of stochastic matrices applied to $\pi_0$ yields a valid probabilistic state vector for any input string.
\end{proof}

\subsection{Proof of Lemma~\ref{LEM:PROBABILISTIC_EPSILON_CLOSURE}: Probabilistic $\varepsilon$-Closure via Matrix Iteration}
\label{PROOF:LEM:PROBABILISTIC_EPSILON_CLOSURE}
\begin{proof}[Proof sketch]
Let $\boldsymbol{p}_t \in [0, 1]^{1 \times n}$ be a row probability vector representing the current state distribution at time $t$, where $[\boldsymbol{p}_t]_i$ denotes the probability of being in state $q_i$. Define the probabilistic $\varepsilon$-transition matrix $T^{\varepsilon} \in [0,1]^{n \times n}$ such that $T^{\varepsilon}_{ij}$ denotes the probability of transitioning from state $q_i$ to $q_j$ via an $\varepsilon$-transition. We assume $T^{\varepsilon}$ is \emph{row-substochastic}: $\sum_{j=1}^n T^{\varepsilon}_{ij} \leq 1$ for each $i$.

Define the iterative sequence:
\[
\boldsymbol{p}^{(k+1)} = \boldsymbol{p}^{(k)} T^{\varepsilon}, \quad \boldsymbol{p}^{(0)} = \boldsymbol{p}_t.
\]

\textbf{Monotonicity:} Since $T^{\varepsilon}$ has nonnegative entries, the sequence $\boldsymbol{p}^{(k)}$ is elementwise non-decreasing:
\[
\boldsymbol{p}^{(k+1)} \geq \boldsymbol{p}^{(k)}.
\]

\textbf{Boundedness:} Because $\boldsymbol{p}^{(0)} \in [0,1]^{1 \times n}$ and $T^{\varepsilon}$ is row-substochastic, the sequence remains bounded in $[0,1]^{1 \times n}$.

\textbf{Convergence:} The sequence $\{\boldsymbol{p}^{(k)}\}$ is monotonic and bounded, so it converges elementwise to a fixed point $\boldsymbol{p}^* \in [0,1]^{1 \times n}$. Furthermore, since each step propagates probability mass only to previously inactive states, and there are at most $n$ states, convergence occurs in at most $n$ steps.

\textbf{Fixed Point Property:} At convergence, we have:
\[
\boldsymbol{p}^* = \boldsymbol{p}^* T^{\varepsilon},
\]
which is a fixed point of the update rule. This vector encodes all probability mass reachable from $\boldsymbol{p}_t$ via any number of $\varepsilon$-transitions.
\end{proof}

\subsection{Proof of Proposition~\ref{PROP:SIMULATION_OF_PFA}: Simulation of PFAs via Feedforward Networks}
\label{PROOF:PROP:SIMULATION_OF_PFA}
\begin{proof}[Proof sketch]
Let the initial state vector be $s_0 := \pi_0 \in \Delta^n$, and apply the finite $\varepsilon$-closure:
\[
s_0 := \pi_0 \Big( \sum_{m=0}^{n-1} (T^\varepsilon)^m \Big).
\]
For each symbol $x_t$ in the input sequence $x = x_1 \cdots x_L$, define the recurrence
\[
s_t = s_{t-1} \, T^{x_t} \Big( \sum_{m=0}^{n-1} (T^\varepsilon)^m \Big), \quad \text{for } t = 1, \dots, L.
\]
By Proposition~\ref{PROP:PROBABILISTIC_STATE_VECTOR_REPRESENTATION}, Lemma~\ref{LEM:PROBABILISTIC_EPSILON_CLOSURE}, and Theorem~\ref{THM:PROBABILISTIC_SUBSET_CONSTRUCTION}, each $s_t \in \Delta^n$ remains a valid probabilistic state vector encoding the distribution after consuming the first $t$ symbols and all intervening $\varepsilon$-steps. Finally, define the network output as
\[
f_\theta(x) = \langle s_L, \mathbf{1}_F \rangle = \sum_{i \in F} [s_L]_i,
\]
which equals the acceptance probability of $\mathcal{A}$ on $x$. Since each operation is a symbolic linear transformation with finite $\varepsilon$-closure, the computation for any length-$L$ input induces a depth-$L$ feedforward \emph{computation graph} of width $n$.
\end{proof}

\subsection{Proof of Proposition~\ref{PROP:PARAMETER_EFFICIENCY}: Parameter Efficiency of Feedforward Automata Simulators}
\label{PROOF:PROP:PARAMETER_EFFICIENCY}
\begin{proof}[Proof sketch]
Each symbol $x \in \Sigma$ is associated with a transition matrix $T^x \in \mathbb{R}^{n \times n}$, where $T^x_{ij} \in [0,1]$ and $\sum_j T^x_{ij} = 1$. Each such matrix contains $n^2$ real-valued entries, which may either be stored directly or parameterized via row-wise softmax normalization. In addition, there is a single $\varepsilon$-transition matrix $T^\varepsilon \in \mathbb{R}^{n \times n}$, which also contributes $n^2$ parameters. The finite closure operator $(T^\varepsilon)^* = \sum_{m=0}^{n-1} (T^\varepsilon)^m$ is computed symbolically and introduces no additional parameters.

Thus, storing all $k$ symbol matrices plus the $\varepsilon$-matrix requires $(k+1)\cdot n^2$ real parameters. The only other components of the model are:
\begin{itemize}[itemsep=0pt, topsep=0pt]
    \item The initial state vector $\pi_0 \in \mathbb{R}^n$, which adds $n$ parameters,
    \item The indicator vector $\mathbf{1}_F \in \{0,1\}^n$, which is fixed and not learned.
\end{itemize}
Hence, the total number of trainable parameters is $O(kn^2)$, regardless of the input length $L$, since the same transition matrices (including $T^\varepsilon$) are reused at each layer of the feedforward network.
\end{proof}

\subsection{Proof of Theorem~\ref{THM:EQUIVALENCE}: Equivalence Between Feedforward Networks and PFAs}
\label{PROOF:THM:EQUIVALENCE}
\begin{proof}[Proof sketch]
We prove both directions constructively.

\textbf{Forward Direction:} Let $\mathcal{A} = (Q, \Sigma \cup \{\varepsilon\}, \delta, q_0, F)$ be a PFA with $|Q| = n$. For each symbol $x \in \Sigma$, let $T^x \in [0,1]^{n \times n}$ be its transition matrix, where each row $i$ defines a distribution $\sum_j (T^x)_{ij} = 1$. Let $T^\varepsilon \in [0,1]^{n \times n}$ denote the $\varepsilon$-transition matrix, and define the finite closure operator $(T^\varepsilon)^* = \sum_{m=0}^{n-1} (T^\varepsilon)^m$.

\textit{Initialization:} Let $s_0 \in \mathbb{R}^{1 \times n}$ be the one-hot \emph{row} vector with $[s_0]_i = \mathbf{1}[q_i = q_0]$, and apply closure:
\[
s_0 = s_0 (T^\varepsilon)^*.
\]

\textit{Recurrence:} For $t = 1, \dots, L$, define:
\[
s_t = s_{t-1} T^{x_t} (T^\varepsilon)^*.
\]
Each $s_t \in \mathbb{R}^{1 \times n}$ is a probability distribution over states after reading $x_1 \dots x_t$ and applying all intervening $\varepsilon$-steps. This follows because $s_0$ is one-hot, each $T^{x_t}$ is row-stochastic, and $(T^\varepsilon)^*$ preserves normalization.

\textit{Final Output:} Let $1_F \in \{0,1\}^n$ be the indicator vector for accepting states. The network's final output is:
\[
f_\theta(x) = \langle s_L, 1_F \rangle = \sum_{q \in F} [s_L]_q,
\]
which is precisely the acceptance probability of the PFA.

\textbf{Reverse Direction:} Suppose $f_\theta$ is a symbolic feedforward network as described above:
\begin{itemize}[itemsep=0pt, topsep=0pt]
    \item Input-dependent transitions are modeled by selecting from a fixed set of transition matrices $\{T^x\}_{x \in \Sigma}$,
    \item An $\varepsilon$-transition matrix $T^\varepsilon$ is applied via the finite closure operator $(T^\varepsilon)^*$,
    \item The state vector is updated via $s_t = s_{t-1} T^{x_t} (T^\varepsilon)^*$,
    \item Output is computed as $f_\theta(x) = \langle s_L, 1_F \rangle$.
\end{itemize}

We now define a PFA $\mathcal{A}' = (Q, \Sigma \cup \{\varepsilon\}, \delta', q_0, F)$ such that:
\begin{itemize}[itemsep=0pt, topsep=0pt]
    \item $Q = \{q_1, \dots, q_n\}$ is the set of hidden states in the network,
    \item $q_0$ corresponds to the one-hot initial state $s_0$,
    \item For every $x \in \Sigma$, define $\delta'(q_i, x)(q_j) = (T^x)_{ij}$,
    \item For the $\varepsilon$-transition, define $\delta'(q_i, \varepsilon)(q_j) = (T^\varepsilon)_{ij}$.
\end{itemize}

Because each $T^x$ and $T^\varepsilon$ is row-stochastic by construction, this defines a valid probabilistic transition function with finite closure. The forward recursion of this PFA matches the feedforward network’s propagation, and the acceptance probability is computed in the same way via $\langle s_L, 1_F \rangle$.

Thus, $f_\theta(x)$ computes exactly $P_{\mathcal{A}'}(x \text{ is accepted})$.
\end{proof}

\subsection{Proof of Proposition~\ref{PROP:LEARNABILITY}: Learnability of PFAs via Symbolic Feedforward Networks}
\label{PROOF:PROP:LEARNABILITY}
\begin{proof}[Proof sketch]
Let $\mathcal{A} = (Q, \Sigma \cup \{\varepsilon\}, \delta, q_0, F)$ be a PFA with $|Q| = n$ states. We construct a symbolic feedforward network $f_\theta$ as follows:

\textbf{1. Input and Transition Representation.}
Each input symbol $x \in \Sigma$ is associated with a stochastic transition matrix $T^x \in [0,1]^{n \times n}$, satisfying $\sum_{j=1}^n T^x_{ij} = 1$ for all $i$ (row-stochastic). In addition, there is a single $\varepsilon$-transition matrix $T^\varepsilon \in [0,1]^{n \times n}$. These matrices are initialized randomly with normalization and treated as trainable parameters.

\textbf{2. Initial State.}
Let $s_0 = e_{q_0} \in \mathbb{R}^{1 \times n}$ be the one-hot \emph{row} vector encoding the initial state. Apply the finite $\varepsilon$-closure:
\[
s_0 = s_0 (T^\varepsilon)^*, \quad (T^\varepsilon)^* = \sum_{m=0}^{n-1} (T^\varepsilon)^m.
\]

\textbf{3. Probabilistic State Propagation.}
For a given string $x = x_1 x_2 \dots x_L$, we recursively compute the hidden state representations:
\[
s_t = s_{t-1} T^{x_t} (T^\varepsilon)^*, \quad \text{for } t = 1, 2, \dots, L.
\]
By construction, each $s_t$ remains a probability row vector in $\Delta^n \subseteq \mathbb{R}^{1 \times n}$.

\textbf{4. Readout and Acceptance Probability.}
Let $v_F \in \{0,1\}^n$ be the indicator vector for the accepting states $F \subseteq Q$. The acceptance probability is computed as:
\[
f_\theta(x) = \langle s_L, v_F \rangle = \sum_{q_i \in F} [s_L]_i.
\]
This quantity corresponds to the probability mass assigned to accepting states after processing $x$, consistent with the PFA semantics.

\textbf{5. Training Objective.}
Given a labeled dataset $\mathcal{D} = \{(x_i, y_i)\}_{i=1}^m$ with ground-truth acceptance probabilities $y_i = P_\mathcal{A}(x_i)$, we train the network using binary cross-entropy (BCE) loss:
\[
\mathcal{L}(\theta) = -\frac{1}{m} \sum_{i=1}^m \Big( y_i \log f_\theta(x_i) + (1-y_i) \log (1 - f_\theta(x_i)) \Big).
\]

\textbf{6. Optimization.}
The transition matrices $\{T^x\}_{x \in \Sigma}$ and $T^\varepsilon$ are optimized using gradient-based updates:
\[
\theta^{(t+1)} = \theta^{(t)} - \eta \nabla_\theta \mathcal{L}(\theta^{(t)}),
\]
with optional projection after each update to ensure row-stochasticity (e.g., row-wise softmax or Sinkhorn-style normalization).

\textbf{7. Empirical Approximation.}
Because the model class includes exact simulators (Theorem~\ref{THM:EQUIVALENCE}) and the training objective directly penalizes deviations from labels, the empirical risk can be driven arbitrarily low on $\mathcal{D}$ given sufficient optimization and capacity. In particular, when labels are realizable by some $\{T^x, T^\varepsilon\}$, there exist parameters with zero training error.
\end{proof}

\end{document}